\documentclass[10pt, twocolumn]{IEEEtran}

\IEEEoverridecommandlockouts
\usepackage{cite}
\usepackage{amsmath,amssymb,amsfonts}
\usepackage{amsthm}
\usepackage{algorithmic}
\usepackage{graphicx}
\usepackage{textcomp}
\usepackage{xcolor}
\usepackage[]{footmisc}
\usepackage{bbm}

\usepackage{cuted}
\usepackage{stfloats}
\usepackage{mathtools,amssymb,lipsum, nccmath}
\newtheorem{theorem}{Theorem}

\usepackage{algorithm,algorithmic}
\usepackage{color,soul}
\usepackage{graphicx}
\usepackage{mathrsfs}
\usepackage{aligned-overset}
\usepackage{subcaption}

\begin{document}

\title{Biased Over-the-Air Federated Learning\\ under
Wireless Heterogeneity}
\author{Muhammad Faraz Ul Abrar, and Nicol\`{o} Michelusi~\IEEEmembership{Senior Member, IEEE}
\thanks{M. Faraz Ul Abrar and N. Michelusi are with the School of Electrical, Computer and Energy
Engineering, Arizona State University. email: \{mulabrar,
nicolo.michelusi\}@asu.edu.
This research has been funded in part by NSF under grant CNS-$2129615$.}
}


\maketitle
\thispagestyle{empty}
\pagestyle{empty}
\setulcolor{red}
\setul{red}{2pt}
\setstcolor{red}


\begin{abstract}
Recently, Over-the-Air (OTA) computation has emerged as a promising federated learning (FL) paradigm that leverages the waveform superposition properties of the wireless channel to realize fast model updates. Prior work focused on the OTA device ``pre-scaler" design under \emph{homogeneous} wireless conditions, in which devices experience the same average path loss, resulting in zero-bias solutions. Yet, zero-bias designs are limited by the device with the worst average path loss and hence may perform poorly in \emph{heterogeneous} wireless settings. In this scenario, there may be a benefit in designing \emph{biased} solutions, in exchange for a lower variance in the model updates. To optimize this trade-off, we study the design of OTA device pre-scalers by focusing on the OTA-FL convergence. We derive an upper bound on the model ``optimality error", which explicitly captures the effect of bias and variance in terms of the choice of the pre-scalers. Based on this bound, we identify two solutions of interest: minimum noise variance, and minimum noise variance zero-bias solutions. Numerical evaluations show that using OTA device pre-scalers that minimize the variance of FL updates, while allowing a small bias, can provide high gains over existing schemes.
\end{abstract}



\section{Introduction}
The unprecedented data availability at the Internet-of-Things (IoT) devices along with their increased computational capabilities has recently shifted the focus from classical machine learning (ML) to distributed learning. Among the distributed learning solutions, FL has gained wide popularity due to its robust privacy guarantees and reduced communication overhead \cite{FL_survey}. A standard FL setting comprises a set of
$N$ devices with their private data collaborating with a central parameter server (PS) e.g., a cloud or edge server, by only sharing their local parameter or gradient information \cite{FL_intro,MH_FL,DistL_intro,FL_survey}.
Typically, the goal is to learn a global FL model parameter
\begin{equation}
\mathbf{w}^* = \arg\min_{\mathbf{w} \in \mathbb{R}^d} F(\mathbf{w}) \triangleq \frac{1}{N} \sum_{m \in [N]} f_m(\mathbf{w}), \tag{P}
\label{FL_prob}
\end{equation}
where $f_m(\mathbf{w})$ represents the local objective function of device $m$, and $F(\mathbf{w})$ is the global objective (loss) function. Typically, \eqref{FL_prob} is solved via iterative algorithms, e.g., mini-batch gradient descent (GD), in which the devices compute local gradients using their datasets, and upload them wirelessly to the PS. Next, the PS \textit{aggregates} the received local gradients, updates the global model and broadcasts it to the devices to complete one FL round.
This process is iterated over several rounds until the global loss function converges\cite{Fedavg}. 

Yet, to realize real-world FL solutions, several practical issues need to be addressed. 
In such systems, numerous low-powered devices need to transmit their local gradient information over a shared
wireless fading channel, necessitating the development of communication-efficient FL schemes \cite{Distl_wireless}. To address this challenge, \cite{OTA_FL,FL_fading,One_bit_FL,Blind_FL,10279097} proposed schemes to perform FL over wireless networks that are robust to channel fading. Another line of work \cite{Sched_policies,ADFL,BB_FL} focused instead on the design of FL device scheduling schemes by taking into account the wireless conditions of the devices.

Recently, OTA computation has emerged as a promising candidate over conventional digital communication approaches for realizing FL solutions that are device-scalable \cite{OTA_FL, FL_fading, BB_FL}. It leverages the fact that concurrent transmissions over wireless multiple access channels (MAC) are superimposed at the receiver \cite{Analog_WSN}. 
A typical requirement for a successful OTA computation scheme is to ensure unbiased OTA aggregation at the receiver, i.e. the received signals should be \emph{aligned} and \emph{equally scaled}, attained by designing OTA ``pre-scalers" and ``post-scaler". Achieving unbiased OTA aggregation over a fading MAC typically requires each device to perform channel inversion, making the choice of the pre-scalers limited by the device with the worst channel conditions. This design may result in a high variance of the FL updates \cite{OTA_FL,BB_FL}, and hence in a deterioration of the convergence performance. To address this limitation,  several works \cite{FL_fading, BB_FL} have proposed thresholding schemes. Nevertheless, prior OTA-FL works in  \cite{FL_fading,OTA_FL,One_bit_FL, OTA_FL_H_data} assume that the devices in the network have the same average path loss, ensuring zero average bias, which is used to provide FL convergence guarantees in \cite{OTA_FL_H_data,One_bit_FL}.

In this paper, we consider a more practical ``wireless heterogeneous" OTA-FL scenario in which the devices may experience different average path losses. It is worth mentioning that using the schemes proposed in \cite{OTA_FL,FL_fading,One_bit_FL,OTA_FL_H_data} under wireless heterogeneity can give rise to FL \textit{objective inconsistency} \cite{Obj_incons} due to non-uniform (biased) device participation, which necessitates studying the impact of this bias on the OTA-FL convergence. It should further be noted that this issue has not been addressed in these works since they have considered homogeneous wireless settings. While threshold-based device scheduling has been proposed to address wireless heterogeneity in \cite{BB_FL}, the impact of the bias on the FL convergence has not been discussed. We address this gap by analyzing the convergence of wireless heterogeneous OTA-FL and derive an upper bound on the expected error in the FL updates, which explicitly captures its dependence on the bias and variance terms. Furthermore, in contrast to \cite{OTA_FL,FL_fading,BB_FL,One_bit_FL, OTA_FL_H_data}, which require the acquisition of global instantaneous channel state information (CSI) to design OTA pre-scalers, here we focus on communication-efficient solutions requiring only statistical CSI. Based on the derived upper bound, we also investigate two interesting OTA-FL device pre-scaler designs: 1) minimum noise variance, 2) minimum noise variance zero-bias pre-scalers. Finally, we numerically demonstrate that under heterogeneous wireless settings, the proposed minimum noise variance \emph{biased} pre-scalers design yields significantly lower global loss and higher test accuracy than existing schemes. 

\textit{Notation}: The space of $n$-dimensional real numbers is denoted by $\mathbb{R}^n$. A boldface lower-case letter represents a vector. A zero mean circularly-symmetric complex Gaussian distributed random variable with variance $\sigma^2$ is denoted by $ \mathcal{CN}$(0, $\sigma^2$). The norm $\Vert \cdot \Vert$ is the Euclidean $\ell$-2 norm. The discrete set $i \in \{1,2,\cdots,N\}$ is denoted by $i \in [N]$, and the expectation of a random variable over the associated probability distribution is denoted by $\mathbb{E}[\cdot]$.



\section{System Model and over-the-air FL}
\begin{figure}[t!]
	\centering\vspace{1mm}
	\includegraphics[width=0.32\textwidth]{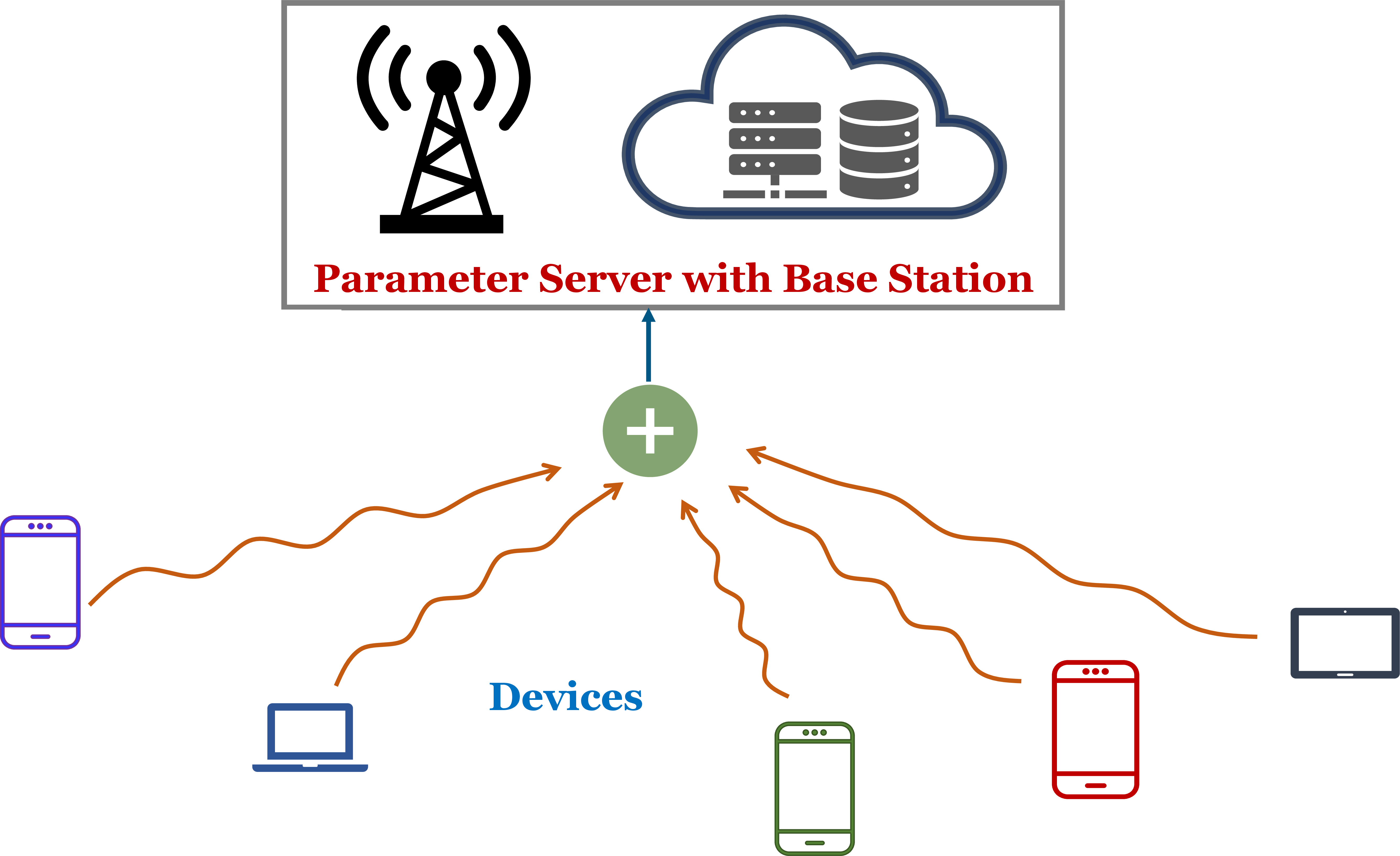}
	\caption{Illustration of OTA-FL system model}
\label{System_model}
\vspace{-5mm}
\end{figure}

We consider a wireless network of $N$ distributed devices coordinating with a base station that also acts as the PS to learn a global model parameter as shown in Fig. \ref{System_model}. The $m$-th device owns a private dataset $\mathcal{D}_m = \{(\boldsymbol{x}^{(1)}_{m}, y^{(1)}_{m}), (\boldsymbol{x}^{(2)}_{m},y^{(2)}_{m}), \cdots \}$, where $\boldsymbol{x}^{(i)}_{m}$ and $y^{(i)}_{m}$ are the feature vector and class label, respectively, associated with the $i$-th local data sample. Each device has a local objective function $f_m(\mathbf{w}) = \frac{1}{\vert \mathcal D_m \vert} \,\sum_{\boldsymbol{\xi} \in \mathcal{D}_m} \phi(\mathbf{w},\boldsymbol{\xi})$, only computable at device $m$, where $\phi(\mathbf{w},\cdot)$ is the loss function, $\boldsymbol{\xi}$ is a data point and $\mathbf{w} \in \mathbb{R}^d$ is the $d$-dimensional learning parameter. We assume a conventional wireless FL setup, in which the solution to \eqref{FL_prob} is obtained by performing GD model updates over multiple FL rounds by aggregating the local gradients. To this end, the FL round $t$ starts with the PS broadcasting the model parameter $\mathbf{w}_t$ to each device. Next, device $m$ uses its full local dataset $\mathcal{D}_m$ to compute the local gradient $\boldsymbol{g}_{m,t} \triangleq \nabla f_m(\mathbf{w}_t)$ 
 at the received parameter $\mathbf{w}_t$ and transmits it to the PS.\footnote{We focus on full-batch GD since it captures the relevant structural aspects of the problem. The extension to stochastic GD is straightforward.} 
 Ideally, the PS aims to
compute the global gradient $\overline{\boldsymbol{g}}_t$, obtained by
aggregating the received local gradients from each device without any errors,
 \begin{align}
\overline{\boldsymbol{g}}_t = \frac{1}{N}\sum_{m \in [N]} \boldsymbol{g}_{m,t}.
\label{Gradagg}
\end{align}
This step is followed by the global model update
 $\mathbf{w}_{t+1}$ as
 \begin{align}
	    \mathbf{w}_{t+1} = \mathbf{w}_{t} - {\eta} \overline{\boldsymbol{g}}_t, \label{GD}
\end{align}
where $\eta$ is the learning stepsize. This process is iterated until the desired accuracy is achieved.
Yet, computing the global gradient in \eqref{Gradagg} requires noiseless aggregation of all the local gradients, i.e., they need to be perfectly aggregated with the desired weight $\frac{1}{N}$ at the PS. However, in practice, the PS instead computes a noisy estimate of the global gradient, $\hat{\boldsymbol{g}}_t$, constructed using local gradients obtained through a wireless channel. Next, we discuss the construction of $\hat{\boldsymbol{g}}_t$ at the PS.

\subsection{Over-the-air transmission over a fading MAC}
In a practical FL system, the transmission of the local gradients $\boldsymbol{g}_{m,t}$ occurs over noisy wireless channels. We model the wireless channel between the devices and the PS as a Rayleigh flat fading channel $h_{m,t} \sim \mathcal{C N} \left(0,\Lambda_m\right) \forall m \in [N],$ i.i.d. over time $t$. Here, $\Lambda_m$ represents the average path loss and is assumed to remain constant during FL running time. Notably, while existing works \cite{OTA_FL,FL_fading,OTA_FL_H_data, One_bit_FL} assume the average path loss to be the same across the devices ($\Lambda_m = \Lambda_n ,\forall m, n \in [N]$), here we assume that it may differ across devices. We also assume that the average path loss knowledge is available at the PS, but not the instantaneous CSI.

We use OTA computation for local gradient transmission, proposed in recent works \cite{OTA_FL,BB_FL,OTA_FL_H_data}. The key idea is to perform joint computation and communication \cite{Analog_WSN}, allowing ``one-shot" local gradient aggregation to realize fast FL updates. To transmit the local gradient, each device $m$, synchronized in time, pre-scales its signal and sends it over a fading uplink MAC to the PS. Let $\mathbf{x}_{m,t}$ denote the signal transmitted by device $m$ in FL round $t$, then the received signal $\mathbf{y}_t$ at the PS can be expressed as
\begin{align}
\mathbf{y}_t = \sum_{m \in [N]} h_{m,t} \cdot \mathbf{x}_{m,t} \;+    
\;\mathbf{z}_t,
\label{Signal_model}
\end{align}
where 
$\mathbf{z}_t \sim \mathcal{C N}(\mathbf{0},N_0 \mathbf{I})$
represents the additive white noise at the PS, i.i.d. over $t$. To approximate the ideal gradient aggregation \eqref{Gradagg} through the signal model \eqref{Signal_model}, we let each device use an OTA pre-scaler $\gamma_m$ and perform a truncated channel inversion. Accordingly, the transmission signal $\mathbf{x}_{m,t}$ is  defined as
\begin{align}
\mathbf{x}_{m,t} =  
\begin{cases}
\frac{\gamma_{m}} { {h}_{m,t}} \boldsymbol{g}_{m,t}, \,\text{ if } \gamma_{m} \leq \sqrt{d E_s} \frac{\vert h_{m,t}\vert}{G_\text{max}},\\
\mathbf 0,\qquad \qquad \,\,\text{otherwise},
\label{signal_transmission}
\end{cases}
\end{align}
where 
$E_s$  is the average energy per sample.
Here, to reduce signaling overhead, we assume that the norm of the local gradients at each round is uniformly bounded, i.e., $\Vert\boldsymbol{g}_{m,t}\Vert \leq G_\text{max}, \, \forall m \in [N], \forall t$ (as also assumed in \cite{niid_fedavg,OTA_FL_H_data}), and $\gamma_m$ remains fixed throughout FL training. Hence, a device does not participate in a round if $\gamma_{m} > \sqrt{d E_s} \frac{\vert h_{m,t}\vert}{G_\text{max}}$, which
ensures the energy constraint $\Vert \mathbf{x}_{m,t}\Vert^2/d \leq E_s\,,\forall m,t$. With this choice of the transmit signal, the PS estimates the global gradient \eqref{Gradagg} as
$\hat{\boldsymbol{g}}_t = \mathbf{y}_t/\alpha$. Using
\eqref{Signal_model} and \eqref{signal_transmission}, it specializes as
\begin{align}
\hat{\boldsymbol{g}}_t = \frac{\mathbf{y}_t} {\alpha}
=\frac{1}{\alpha} \sum_{m \in [N]}\chi_m\gamma_{m}\boldsymbol{g}_{m,t}
\;+    
\;\frac{\mathbf{z}_t}{\alpha},
\label{Signal_model2}
\end{align}
where $\chi_m$ is the indicator of 
the transmit decision in \eqref{signal_transmission}, and 
$\alpha$ is the OTA post-scaler.
Due to concurrent uplink transmissions by the devices, the overall gradient upload time in each round $t$ is $\frac{d}{B}$, where $B$ denotes the bandwidth shared by the devices.
\subsection{Biased Over-the-Air-FL}
The PS then updates the global model using \eqref{Signal_model2} as
\begin{align}
 \mathbf{w}_{t+1} = \mathbf{w}_{t} - \eta \hat{\boldsymbol{g}}_t. \label{Prac_GD}
\end{align}
It is straightforward to verify that $\mathbb{E} [\mathbf{y}_t ] = \sum_{m \in [N]} \alpha_{m} \boldsymbol{g}_{m,t}$, where $\alpha_{m} =  {\gamma_{m} e^ \frac{ - \gamma_{m}^2 G_\text{max}^2}{d \Lambda_m  E_s}}$ and the expectation is over channel fading and white noise at the PS, conditioned on $\mathbf{w}_t$. The PS designs the post-scaler as $\alpha = \sum_{m \in [N]} \alpha_{m}$. With this choice, the expected estimate of the global gradient $\tilde{\boldsymbol{g}}_t \triangleq \mathbb{E}[\hat{\boldsymbol{g}}_t]$ is a convex combination of the local gradients of the devices, i.e.,

\begin{align}
\tilde{\boldsymbol{g}}_t = \sum_{m \in [N]} p_m \boldsymbol{g}_{m,t}\,,\label{exp_global_grad} 
\end{align}
where $p_m \triangleq \frac{\alpha_m}{\alpha}$ can be interpreted as the OTA-FL average \textit{participation level} of device $m$, where $0{\leq}p_m{\leq}1, \sum_{m \in [N]} p_{m} = 1$. 
 With this definition, note that \eqref{Prac_GD} is a noisy (stochastic) gradient descent algorithm, which, on average, updates the global FL model using $\tilde{\boldsymbol{g}}_t$ as in \eqref{exp_global_grad}, in place of 
 $\bar{\boldsymbol{g}}_t$ in \eqref{Gradagg}. 
 Therefore, these updates minimize a different objective function than \eqref{FL_prob}, on average, given by
 \begin{align}
    \tilde{F}(\mathbf{w}) = \sum_{m \in [N]} p_m f_m(\mathbf{w}).
    \tag{$\tilde{\text{P}}$}
    \label{incons_obj}
\end{align}
This can be seen by noting that $\tilde{\boldsymbol{g}}_t = \mathbb{E}[\hat{\boldsymbol{g}}_t]$ is the gradient of $\tilde{F}$ at $\mathbf w_t$.
We highlight here that the existing schemes \cite{FL_fading,OTA_FL,One_bit_FL, OTA_FL_H_data} assume the same average path loss across devices, yielding uniform device participation, $p_m= \frac{1}{N}, \forall m \in [N]$, so that \eqref{incons_obj} and \eqref{FL_prob} become equivalent. However, these schemes, when used in a heterogeneous wireless setting, minimize a different objective function causing the issue of objective inconsistency \cite{Obj_incons}, and introducing a model bias. Consequently, their convergence guarantees do not apply to the wireless heterogeneous setting studied in this paper. On the other hand, while forcing zero bias performs well under homogeneous wireless settings, see e.g. \cite{OTA_FL_H_data}, it may yield high variance in FL updates under heterogeneous wireless conditions, motivating a biased OTA-FL design studied in this paper. Let $\tilde{\mathbf{w}}$ denote the solution to $\min\limits_{\mathbf{w} \in \mathbb{R}^d} \tilde{F}(\mathbf{w})$. 
 In the next section,
 we characterize the associated model bias $\Vert\tilde{\mathbf{w}} -  \mathbf{w}^* \Vert$, and study its impact on the convergence of OTA-FL.

\section{Convergence Analysis and pre-scaler design}
In this section, we theoretically characterize the learning performance of a biased OTA-FL system as described previously in terms of the choice of the OTA device pre-scalers. We analyze the convergence to the global minimum of $\eqref{FL_prob}$ using the metric $\sqrt{\mathbb{E}\left[\Vert \mathbf{w}_{t} - \mathbf{w}^* \Vert^2 \right]}$, which we refer to as the model ``optimality error". It measures the expected deviation of the current model $\mathbf{w}_{t}$ from the global minimizer $\mathbf{w}^*$.
To study the convergence, we require the following assumptions:\\
\textbf{{Assumption 1}}. Each local objective function $f_m(\mathbf x)$ is $L_m$-smooth and $\mu_m$-strongly convex. It follows that $F(\mathbf{w})$ and $\tilde{F}(\mathbf{w})$ are $L$ and $\tilde{L}$ smooth and $\mu$ and $\tilde{\mu}$-strongly convex respectively, where $L = \frac{1}{N} \sum_{m \in [N]} L_m$, $\tilde{L} = \sum_{m \in [N]} p_m L_m$, $\mu = \frac{1}{N} \sum_{m \in [N]} \mu_m$, and $\tilde{\mu} = \sum_{m \in [N]} p_m \mu_m$. \\
\textbf{{Assumption 2}}.  The average of the squared norm of the local gradients at the global minimizer $\mathbf{w}^*$ is bounded, i.e.,
$\frac{1}{N}\sum_{m \in [N]} \left\| \nabla f_m(\mathbf{w}^*)\right\|^2 \leq  \kappa^2$; $ \kappa =0$ corresponds to the case when the local objectives $f_m$ are identical across devices.
\textbf{{Assumption 3}}. The norm of local gradients in each FL round is uniformly bounded, i.e., $\Vert\boldsymbol{g}_{m,t}\Vert \leq G_\text{max}, \, \forall m \in [N], \forall t$.

Note that Assumption 1 is standard in the literature used to study FL convergence. Assumption 2 is weaker than the assumption of bounded local gradient dissimilarity in \cite{Obj_incons}, and Assumption 3 has also been used in \cite{niid_fedavg,OTA_FL_H_data}.
\subsection{Main Convergence Results}
Now, we are ready to present our main convergence result. Since the iterative algorithm described in \eqref{Prac_GD} on average minimizes \eqref{incons_obj}, we approach the analysis by splitting the overall error into: the error between $\mathbf w_t$ and $\tilde{\mathbf{w}}$ (the minimizer  of \eqref{incons_obj});
the error between $\tilde{\mathbf{w}}$ and the global minimizer $\mathbf{w}^*$. Define $\|\mathbf{w}_{t} - \mathbf{w}^*\|^2 \triangleq E_{t}$, and $\|\mathbf{w}_{t} - \tilde{\mathbf{w}}\|^2 \triangleq \Tilde{E}_{t}$, then the optimality error can be upper bounded as shown next.
\begin{theorem}
\label{thm1}
With local objective functions $f_m(\mathbf{w})$ satisfying Assumptions 1-3, and fixed learning stepsize $\eta \in [0, \frac{2}{\tilde{\mu} + \tilde{L}}]$, the optimality error given $E_0$ after $t$ FL rounds satisfies
\begin{multline}
\sqrt{\mathbb{E}[E_{t}]} \leq \underbrace{
\left(1-\eta \tilde{\mu}\right)^{t}\sqrt{\tilde{E}_{0}}}_\text{initialization error} +  \underbrace{ \frac{N \kappa}{\tilde{\mu}} \max_{m \in [N]} \left|\frac{1}{N} -p_m\right|}_\text{model bias} + \\\Bigg( \frac{\eta}{\tilde{\mu}} \Big(\underbrace{\sum_{m \in [N]} p_{m}^2 G^2_\text{max} \Big(\frac{\displaystyle \gamma_{m} }{\displaystyle \alpha_{m}} - 1 \Big)}_\text{transmission variance} +  \underbrace{\frac{\displaystyle d N_0}{\displaystyle \alpha^2}}_\text{noise variance}\Big)\Bigg)^{1/2}.
\label{UB}
\end{multline}
\end{theorem}
The proof sketch of Theorem \ref{thm1} is provided in the Appendix.

Note that the expression derived in \eqref{UB} explicitly shows the convergence behavior of the biased OTA-FL in terms of four key terms: 1) FL initialization 2) model bias 3) transmission variance 4) noise variance. We highlight that the model bias term arises mainly from the fact that we have considered arbitrary device participation levels $p_m$, and hence a zero bias is achievable only with either uniform device participation ($p_m = 1/N, \forall m \in [N])$ or identical objective functions ($\kappa= 0$). The transmission variance results from the intermittent transmission of the local gradients. To elaborate, due to fluctuations in channel realizations $h_{m,t}$ at each iteration, for a given choice of $\gamma_m$, a device is only able to upload its local gradient according to \eqref{signal_transmission} while satisfying the energy budget. Finally, the noise variance term arises because the updates are affected by the noise at the PS. We note here that, while the transmission variance can be reduced by choosing smaller values for $\{\gamma_m\}$, such a design causes high noise variance. On the other hand, reducing the impact of noise variance can lead to a non-zero model bias. Thus, the problem of choosing OTA device pre-scalers for improved FL performance is worth addressing.
\subsection{OTA pre-scalers design}
To design the device pre-scalers, we consider the problem:
\begin{align}
    \min\limits_{\{\gamma_m\},\gamma_m > 0\,, m\in [N]}  \Psi(\{\gamma_m\}),
    \label{OTA_design} \tag{P1}
\end{align}
where we define $\Psi(\{\gamma_m\})$ as the upper bound on $\sqrt{\mathbb{E}[E_{t}]}$ in \eqref{UB}. Note that \eqref{OTA_design} is a non-convex optimization problem due to the model bias and the square root of the sum of the variance terms being non-convex in $\gamma_m$. Nevertheless, we would like to mention here that in a practical FL setting, the noise variance term in \eqref{UB} is typically the major bottleneck. Therefore, we provide here two interesting solutions that minimize the two key terms in $\Psi(\{\gamma_m\})$: 
1) minimum noise variance solution, and
2) (minimum variance) zero-bias solution. The considered design of pre-scalers will remain fixed throughout FL training.\\
1) \underline{Minimum noise variance solution}: To minimize the term $\frac{dN_0}{\alpha^2}$, it is obvious to choose $\{\gamma_m\}_{m=1}^N$ which maximizes $\alpha$. Note that $\alpha = \sum_{m \in [N]}{\gamma_{m} e^ \frac{ - \gamma_{m}^2 G_\text{max}^2}{d \Lambda_m  E_s}} $,
and ${\gamma_{m} e^ \frac{ - \gamma_{m}^2 G_\text{max}^2}{d \Lambda_m  E_s}}$ is log-concave in $\gamma_{m}$. Thus, it can be verified that $\{\gamma_m\}_{m=1}^N$ that minimizes the noise variance is given by
\begin{align}
\Tilde{\gamma}_m = \sqrt{\frac{d \Lambda_m E_s}{2 G_\text{max}^2}}, \quad \forall m \in [N].
\end{align}
2) \underline{Zero-bias solution}: It requires minimizing the bias term $\frac{N \kappa}{\tilde{\mu}} \max\limits_{m \in [N]} \left|\frac{1}{N} -p_m\right|$, which can be made zero for a family of solutions of $\{\gamma_m\}_{m=1}^N$ that guarantees uniform expected device participation. Among these solutions, here we discuss a zero-bias solution that minimizes the noise variance, which we denote by $\{\Bar{\gamma}_m\}$.
Note that any zero-bias solution requires $\alpha_m={\gamma_{m} e^ \frac{ - \gamma_{m}^2 G_\text{max}^2}{d \Lambda_m  E_s}}=\alpha/N,\forall m$. Further, observe that $\alpha_m \leq \alpha_m(\tilde{\gamma}_m), \forall m$, where $\Tilde{\gamma}_m$ is the minimum noise variance pre-scaler. Without loss of generality, assume the devices are ordered such that, $\Lambda_1 \geq \Lambda_2 \geq \cdots \geq \Lambda_N$, then it can be easily verified that a zero bias solution with the minimum noise variance (i.e., maximum $\alpha$) is obtained by setting $ \forall m, \alpha_m(\Bar{\gamma}_m) = 
\min_{m'\in[N]}\alpha_{m'}(\tilde{\gamma}_{m'})=
\alpha_N(\tilde{\gamma}_N)$, where the desired solution $\{\Bar{\gamma}_m\}$ can be expressed as a Lambert W function. This choice of pre-scaler results in highest feasible post-scaler, i.e., $\alpha = N \alpha_N(\tilde{\gamma}_N) =  N \alpha_N(\bar{\gamma}_N)$. Finally, note that any $\alpha < N \alpha_N(\bar{\gamma}_N)$ yields higher noise variance, confirming the desired solution.

The provided solutions can also be used to initialize an iterative algorithm e.g., subgradient descent to solve \eqref{OTA_design}. This variant is left for future work.

\section{Numerical Results}
In this section, we perform numerical experimentation to evaluate the performance of our proposed schemes. We study the handwritten digit classification problem in an FL setting on the popular MNIST dataset \cite{MNIST}, which consists of C = 10 classes from ``0” to “9”. We perform softmax regression on a single-layer neural network with each image of size 28 x 28 pixels. We consider the FL problem with $N= 10$ devices uniformly deployed within a radius of $r_\text{max} = 200 $ m from the PS situated at the center. The devices share a bandwidth $B$ = 1 MHz and communicate over a carrier frequency $f_c =$ 2.4 GHz with transmission power $P_\text{tx} =$ 20 dBm. The noise power spectral density at the PS is $N_0 = -174$dBmW/Hz. The average path loss $\Lambda_m$ between the devices and the PS follows the log-distance path loss model with path loss exponent $\beta = 2.2$ and $40$ dB loss at the reference distance of 1 m. The optimization parameter $\mathbf{w} \in \mathbb{R}^{7850}$ is given as $\mathbf{w}^T =$  
$\begin{bmatrix}
{\mathbf{w}^{(0)}}^T,\cdots,{\mathbf{w}^{(9)}}^T \end{bmatrix}$, where $\mathbf{w}^{(\ell)}$ is the sub-parameter associated with class $\ell$. We use the regularized cross-entropy loss function at each device, given by
\begin{align*}
\phi((\boldsymbol{x},\ell );\mathbf{w})= \frac{0.01}{2}\Vert \mathbf{w}\Vert^2 -\ln \left(\frac{\exp{\{ \boldsymbol{x}^T\mathbf{w}^{(\ell)}\}}}{\sum_{c = 0}^9{\exp{\{ \boldsymbol{x}^T\mathbf{w}^{(c)}\}}}}\right),
\end{align*}


\begin{figure*}[t]
\centering
\begin{subfigure}[t]{0.325\textwidth}
    \centering
    \includegraphics[width=\textwidth,trim=18 0 35 15, clip]{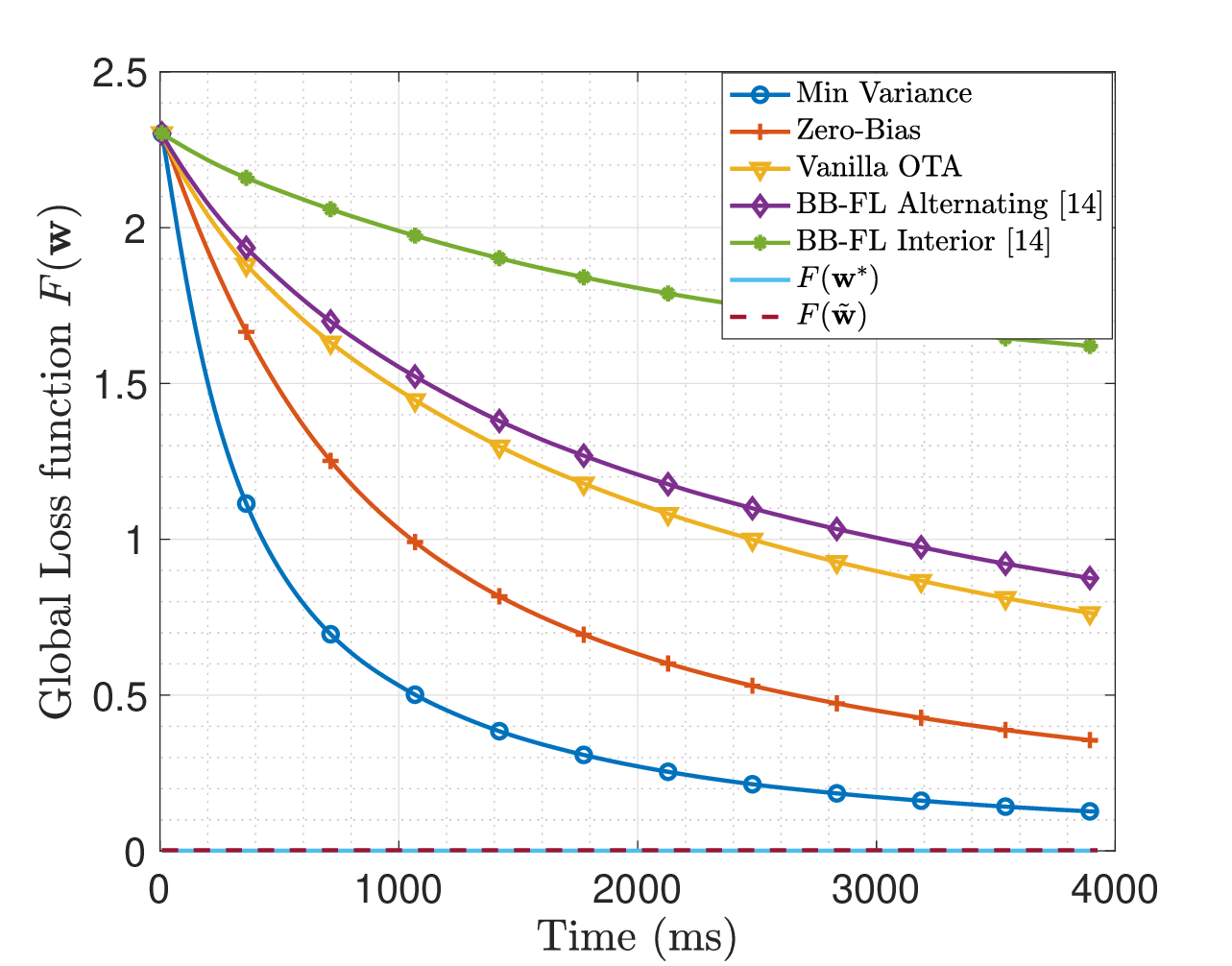}
    \caption{Global objective function $F(\mathbf{w})$ over training time (ms), $N =10$ devices.} \label{fig1a}
\end{subfigure}
\begin{subfigure}[t]{0.325\textwidth}
    \centering
    \includegraphics[width=\textwidth,trim=18 0 35 15, clip]{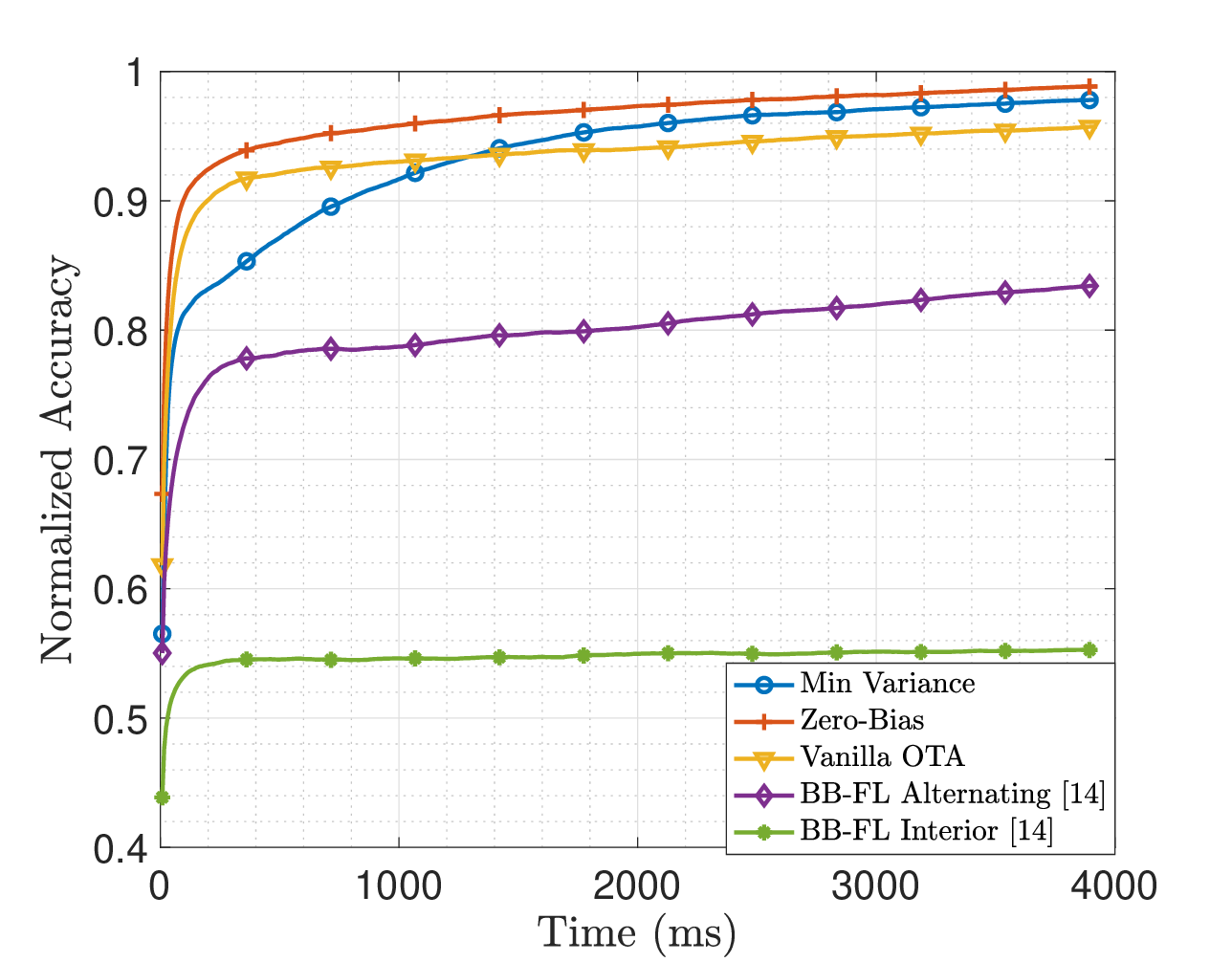}
    \caption{Normalized accuracy over training time (ms), $N =10$ devices.} \label{fig1b}
\end{subfigure}
\begin{subfigure}[t]{0.325\textwidth}
    \centering
    \includegraphics[width=\textwidth,trim=8 0 50 15, clip]{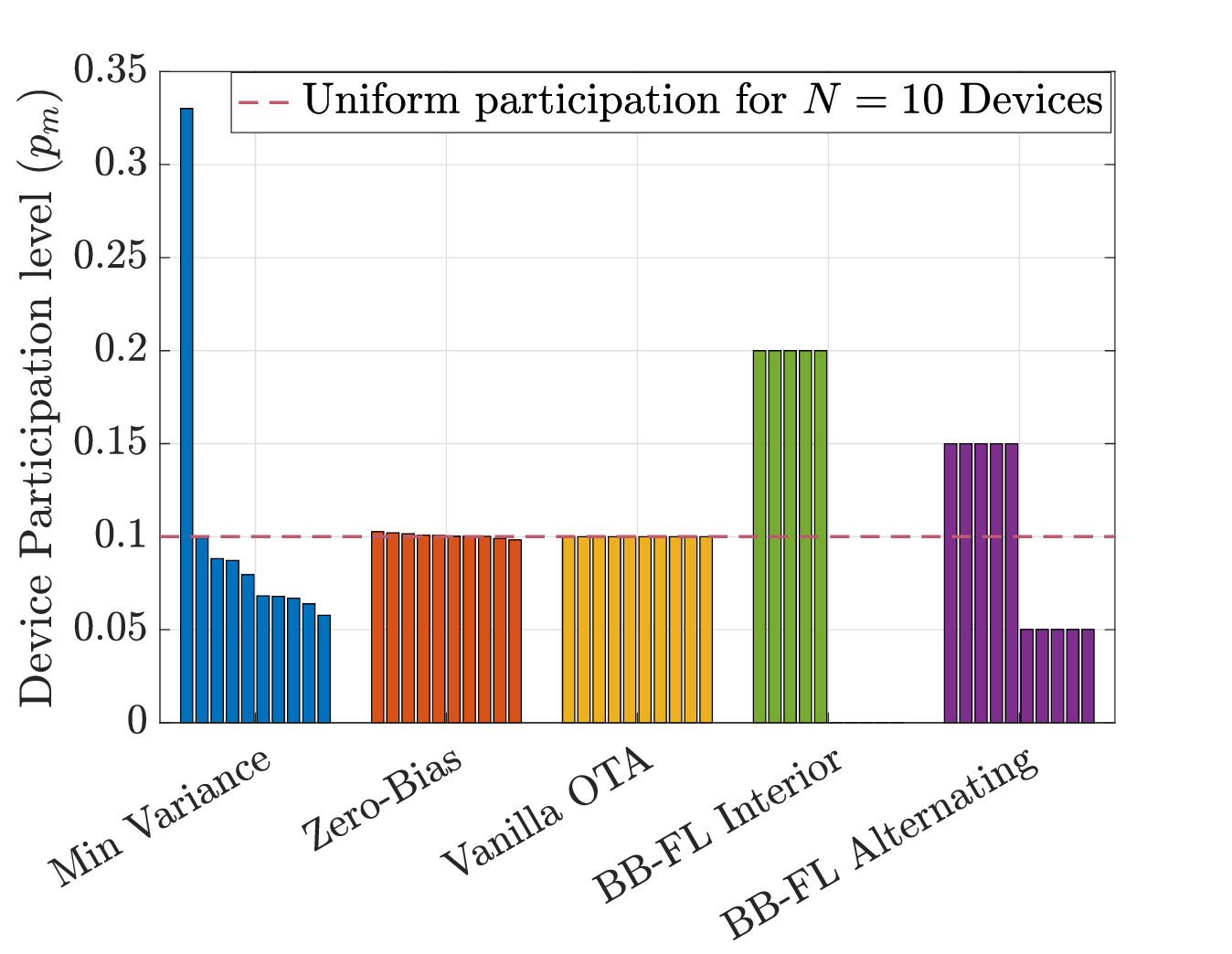}
    \caption{Average device participation level, $N =10$ devices.} \label{fig1c}
\end{subfigure}
\caption{Comparison of various OTA-FL schemes}
\vspace{-2mm}
\label{FL_results}
\end{figure*}
\noindent where we assume $\mu_m = 0.01$ for each device. Since in most practical FL scenarios, devices possess limited, albeit unique, data, we perform experiments with training data with overall $\sum_{m \in [N]} \vert \mathcal{D}_m\vert = 100$ datapoints with 10 samples associated to each class, and realize a non-i.i.d. data deployment (data heterogeneity) well-suited for FL applications. For this, we arbitrarily assign a unique label to each device, such that all the datapoints of that class belong only to one device.

To demonstrate the effectiveness of our analysis, we evaluate the performance of both minimum noise variance (biased) and minimum noise variance zero-bias solutions. In addition, we also make comparisons with several state-of-the-art OTA-FL schemes: 1) Vanilla OTA scheme \cite{OTA_FL}, in which each device uses OTA computation to have zero instantaneous bias in each FL round, 2) BB-FL Interior \cite{BB_FL}, which allows only the devices within a radius $R_\text{in} < r_\text{max}$ to perform OTA aggregation, and 3) BB-FL Alternative \cite{BB_FL}, which alternates randomly between scheduling every device and BB-FL Interior policy. It is worth highlighting that the schemes in \cite{BB_FL} also address the issue of wireless heterogeneity in OTA-FL in a heuristic fashion, making them suitable candidates for comparison. We clarify that while schemes 1-3 require instantaneous CSI, as opposed to the two proposed schemes which only require statistical CSI, we neglect the additional overhead incurred. We set $R_{\text{in}} = 0.6\,r_{\text{max}}$ for BB FL Interior and BB FL Alternative schemes for best performance, as demonstrated in \cite{BB_FL}. Moreover, we have chosen the best constant learning stepsize $\eta$ for each scheme obtained via a grid search. 

In Fig. \ref{FL_results}, we show the performance of the above-mentioned OTA-FL schemes over a training duration of $4000$ ms, for a fixed deployment averaged over channel and noise realizations. Fig. \ref{FL_results}a shows the global loss function $F(\mathbf{w})$ over FL training, whereas we plot normalized test accuracy (with respect to that of the global minimizer $\mathbf{w}^*$) in Fig. \ref{FL_results}b. It can be observed that the best performance in terms of global loss is achieved by the proposed Minimum Variance, followed by Zero-Bias schemes, and then other existing OTA-FL schemes. This is because the Minimum Variance scheme, instead of forcing unbiased updates, assigns a pre-scaler to each device according to its (possibly different) average path loss and hence allows a non-zero bias. Thanks to the reduced noise variance, the Minimum variance scheme exhibits the fastest global loss decay rate. On the other hand, while the proposed Zero-Bias scheme exhibits a slower global loss decay than the Minimum Variance scheme due to relatively higher noise variance, it ensures uniform average device participation and hence asymptotically converges to $\mathbf{w}^*$. As a result, it achieves the best final accuracy of  98\% (of the accuracy at $\mathbf{w}^*$). We further highlight that while the Vanilla OTA scheme also designs the pre-scalers such that the estimate of the global gradient is unbiased in each round, the proposed Zero-Bias scheme is more flexible as it only ensures zero bias on average, thereby performing remarkably better ($\approx$ 2.5$\times$ time reduction for the same global loss). Since each device brings a unique label's samples into the network, among the schemes of \cite{BB_FL}, BB-FL Alternating performs better than BB-FL Interior, which conforms with the findings of \cite{BB_FL}. For the BB-FL Interior, since only a subset of devices participate throughout the FL training, the model is unable to generalize on the samples of the unseen classes resulting in worse performance. Finally, while Vanilla OTA performs well compared to BB-FL schemes, the high noise variance resulting from forcing zero instantaneous bias becomes a bottleneck in achieving faster convergence. 

Fig. \ref{FL_results}c shows the average device participation level for the considered schemes in order of decreasing path losses. Clearly, both the Zero-bias scheme and Vanilla OTA exhibit uniform device participation. On the other hand, the Minimum Variance, BB-FL Interior, and BB-FL Alternating schemes allow unequal device participation. Nevertheless, unlike the latter schemes, which use a heuristic approach for variance reduction for FL updates, the former (proposed) scheme allows non-uniform device participation with the aim of faster OTA-FL convergence. Overall, it can be concluded instead of forcing zero bias in each round, the proposed pre-scalers designs with average unbiasedness, and a small bias, yield almost 2$\times$, and 4$\times$ time reduction to achieve the same accuracy, respectively.

\section{Conclusion}

In this paper, we have studied the performance of an OTA-FL system when devices have heterogeneous wireless conditions. We characterized the performance in terms of convergence behavior and derived an upper bound on the optimality error. Unlike existing works, which force zero-bias FL updates, we studied the convergence allowing biased updates. We have shown through the analysis that in the presence of wireless heterogeneity, the optimality error decomposes into respective bias and variance terms. To prove the efficacy of our analysis for OTA device pre-scaler design, we provide two pre-scalers choices using the derived upper bound. We also performed numerical evaluations to support our analysis. We numerically showed that minimizing the model noise variance results in superior performance over existing schemes in a heterogeneous wireless environment with a negligible bias.

\appendix
\textit{Proof sketch of upper bound on ${E}_{t}$ in \eqref{UB}.} We start by expressing the optimality error in terms of expected error between the model updates $\mathbf{w}_{t+1}$ and the minimizer of $\Tilde{F}(\mathbf{w})$ i.e., $\Tilde{\mathbf{w}}$, and the distance between $\Tilde{\mathbf{w}}$ and ${\mathbf{w}}^*$. Recall, $E_{t+1} = \|\mathbf{w}_{t+1} - \mathbf{w}^*\|^2 $, and $ \Tilde{E}_{t+1} = \|\mathbf{w}_{t+1} - \tilde{\mathbf{w}}\|^2$. By Minkowski's inequality \cite{Minkowski_ref}
and
$\mathbf{w}_{t+1} - \mathbf{w}^*
=
(\mathbf{w}_{t+1}-\tilde{\mathbf{w}})+(\tilde{\mathbf{w}} - \mathbf{w}^*)$,
\begin{align}
\sqrt{\mathbb{E} [E_{t+1}]} &\leq \sqrt{\mathbb{E} [\tilde E_{t+1}]} + \|{\tilde{\mathbf{w}}} - \mathbf{w}^*\|.
\label{Minkowski}
\end{align}
First, we establish an upper bound on the first term of the right-hand side of \eqref{Minkowski}. By the definition of FL model updates in \eqref{Prac_GD}, we have $\tilde{E}_{t+1} = \left\|\mathbf{w}_{t} - \eta \hat{\boldsymbol{g}}_t - \tilde{\mathbf{w}}\right\|^2 $, where $\hat{\boldsymbol{g}}_t$ is the estimate of the global gradient, which can be expressed as
\\\centerline{$\hat{\boldsymbol{g}}_t=\sum_{m \in [N]} p_m \nabla{f}_m(\mathbf{w}_t) + \boldsymbol{e}_t
=
\nabla\tilde{F}(\mathbf{w}_t) 
+\boldsymbol{e}_t
,$}
where $\boldsymbol{e}_t=
\hat{\boldsymbol{g}}_t-\mathbb E[\hat{\boldsymbol{g}}_t|\mathbf w_t]$
is a zero-mean error on the estimate of $\nabla\tilde{F}(\mathbf{w}_t)$, and recall $\tilde{F}(\mathbf{w}_t) = \sum_{m \in [N]} p_m f_m (\mathbf{w}_t)$.
Next, we establish the optimality error conditional on $\mathbf{w}_t$ i.e., $\mathbb{E}[\tilde{E}_{t+1} |\mathbf{w}_t]$. Using
$\mathbf{w}_{t+1}=\mathbf{w}_{t}-\eta\nabla\tilde{F}(\mathbf{w}_t) 
-\eta\boldsymbol{e}_t$
and
$\mathbb{E} [\boldsymbol{e}_t|\mathbf{w}_t]{=}\mathbf 0$, we find
\begin{align}
\mathbb{E}[\tilde{E}_{t+1} \mid \mathbf{w}_t] = &\Tilde{E}_{t} + \eta^2 \Vert\nabla\tilde{F}(\mathbf{w}_t)\Vert^2  - 2\eta \nabla\tilde{F}(\mathbf{w}_t)^T\left( \mathbf{w}_{t} -  \tilde{\mathbf{w}}\right)\nonumber\\&+ \eta^2\mathbb{E}[ \left\|\boldsymbol{e}_t\right\|^2 |\mathbf w_t].
\label{dfghsd}
\end{align}The first three terms of the right-hand side can be thought of as a  sequence of GD updates $\{\mathbf{w}_{t}\}$ that is solving \eqref{incons_obj}. Invoking the $\tilde{\mu}$ strong convexity and $\tilde{L}$ smoothness of $\tilde{F}(\mathbf{w})$ in Assumption 1, we use \cite[Lemma 3.11]{bubeckconvex}. It states that
\begin{align}
\nabla\tilde{F}(\mathbf{w}_t)^T\left( \mathbf{w}_{t} {-}  \tilde{\mathbf{w}}\right) \geq \frac{\tilde{\mu}\tilde{L}}{\tilde{\mu} + \tilde{L}} \tilde{E}_t + \frac{1}{\tilde{\mu} + \tilde{L}} \Vert \nabla\tilde{F}(\mathbf{w}_t)\Vert^2.
\label{bubeck}
\end{align}
We use this bound in \eqref{dfghsd},
under the learning stepsize condition $\eta \in [0, \frac{2}{\tilde{\mu} + \tilde{L}}]$,
 followed by strong convexity, which implies $\Vert \nabla\tilde{F}(\mathbf{w}_t)\Vert^2 \geq \Tilde{\mu}^2 \tilde{E}_{t}$.
These steps yield
\begin{align}
\mathbb{E}[\tilde{E}_{t+1} \mid \mathbf{w}_t] \leq \left(1-\eta \tilde{\mu}\right)^2 \tilde{E}_{t} + \eta^2\mathbb{E}[ \|\boldsymbol{e}_t\|^2 |\mathbf w_t].
\label{one_step}
\end{align}

Now, conditioning on $\mathbf{w}_t$, we proceed to compute $\mathbb{E}[ \|\boldsymbol{e}_t\|^2]$ to describe \eqref{one_step}. Note that $\mathbb{E} [\| \boldsymbol{e}_t\|^2] = \mathbb{E} [\|\hat{\boldsymbol{g}}_t  - \mathbb{E}[\hat{\boldsymbol{g}}_t ]\|^2] = \mathbb{E} [ \Vert\hat{\boldsymbol{g}}_t\Vert^2] - \Vert \mathbb{E}[\hat{\boldsymbol{g}}_t ]\Vert^2$ resulting in,
\begin{multline*}
\mathbb{E}\left[ \left\|\boldsymbol{e}_t\right\|^2 \mid \mathbf{w}_t\right]= \sum_{m \in [N]} p_{m}^2 \Vert\boldsymbol{g}_{m,t}\Vert^2 \left(\frac{\gamma_{m} }{\alpha_{m}} - 1 \right)+  \frac{d N_0}{\alpha^2},
\end{multline*}
where recall $p_m = \frac{\alpha_m}{\alpha}$. Using Assumption 3 on the local gradient norm, we further upper bound $\mathbb{E}[ \|\boldsymbol{e}_t\|^2 \mid \mathbf{w}_t]$ as
\begin{align}
    \mathbb{E}\left[ \left\|\boldsymbol{e}_t\right\|^2 | \mathbf{w}_t\right] \leq \sum_{m \in [N]} p_{m}^2 G_\text{max}^2 \left(\frac{\gamma_{m} }{\alpha_{m}} {-} 1 \right){+}  \frac{d N_0}{\alpha^2}\triangleq \sigma^2.
    \nonumber
\end{align}
Finally, 
we compute the expectation over $\mathbf{w}_t$ and
we use induction and the fact that $\eta \tilde{\mu} \leq 1$ to express the optimality error given $\tilde{E}_0$ as
\begin{align}
\mathbb{E}[\tilde{E}_{t}] \leq (1 - \eta \Tilde{\mu})^{2t} \tilde{E}_0 + \frac{\eta}{\tilde{\mu}} \sigma^2.
\label{t+1_step}
\end{align}
Now, we proceed to establish a bound on the second term on the right-hand side in \eqref{Minkowski}, i.e., on
$\|{\tilde{\mathbf{w}}} - \mathbf{w}^*\|$ capturing the model bias. To this end,  
since $\tilde{F}(\mathbf{w})$ is $\Tilde{\mu}$-strongly convex, it follows that $\tilde{\mu}\|{\tilde{\mathbf{w}}} - \mathbf{w}^*\|{\leq}\| \nabla\tilde{F}(\tilde{\mathbf{w}}) {-} \nabla\tilde{F}({\mathbf{w}^*}) \|  =\|  \nabla\tilde{F}({\mathbf{w}^*}) \| $. 
Furthermore, for arbitrary $\mathbf{w}$, 
$$\| \nabla F(\mathbf{w}) -  \nabla\tilde{F}({\mathbf{w}})\|^2 = \Big\| \sum_{m \in [N]} \Big(\frac{1}{N} -p_m \Big) \nabla f_m(\mathbf{w}) \Big\|^2 $$$${\underset{(a)}{\leq}} 
\sum_{m \in [N]}\Big(\frac{1}{N} -p_m \Big)^2
\sum_{m \in [N]}\| \nabla f_m(\mathbf{w}) \|^2
,$$ where ($a$) uses the triangular inequality, followed by Cauchy–Schwarz inequality.
 By evaluating this bound at the global minimizer $\mathbf{w}^*$ (hence, $\nabla{F}({\mathbf{w}^*})=\mathbf 0$) and using
Assumption~2, we obtain
 \begin{align*}
\|\nabla\tilde{F}({\mathbf{w}}^*) \|^2& {\leq} N \kappa^2 \sum_{m \in [N]} \left(\frac{1}{N} -p_m\right)^2\\& \leq N^2 \kappa^2 \max_{m \in [N]} \left(\frac{1}{N} - p_m\right)^2.\end{align*}
Combining this bound with the previous result $\tilde{\mu}\|{\tilde{\mathbf{w}}}{-}\mathbf{w}^*\|{\leq}\|  \nabla\tilde{F}({\mathbf{w}^*}) \| $, it immediately follows that
\begin{align}
\|{\tilde{\mathbf{w}}} - \mathbf{w}^*\| \leq \frac{N \kappa}{\tilde{\mu}} \max_{m \in [N]} \left\vert\frac{1}{N} -p_m\right\vert. 
 \label{dist_min}
\end{align}
Using \eqref{t+1_step} and \eqref{dist_min} in \eqref{Minkowski} completes the proof.

\bibliographystyle{IEEEtran} 
\bibliography{Refs} 


\end{document}